\documentclass{amia}
\usepackage{amsmath}
\usepackage{booktabs}
\usepackage{caption}
\usepackage{colortbl}
\usepackage{graphicx}
\usepackage{lipsum}  
\usepackage{subcaption}
\usepackage[table]{xcolor}



\begin{document}
\renewcommand{\thefootnote}{\fnsymbol{footnote}} % Change footnote style to symbols
\title{Automated Thematic Analysis for Clinical Qualitative Data: Iterative Codebook Refinement with Full Provenance}
\author{
Seungjun Yi, MS$^{1}$,
Joakim Nguyen, MS$^{2}$,
Huimin Xu, MS$^{3}$,
Terence Lim, PhD$^{4,5}$, \\ 
Joseph Skrovan, MS$^{3}$, 
Mehak Beri, MS$^{3}$, 
Hitakshi Modi, MS$^{6}$, \\ 
Andrew Well, MD, MPH, MSHCT$^{7,8}$,
Carlos M Mery, MD$^{7,8}$, \\
Yan Zhang, PhD$^{3}$,
Mia K Markey, PhD$^{1}$,
Ying Ding, PhD$^{3,11}$\thanks{Bill \& Lewis Suit Professor, School of Information, The University of Texas at Austin.}
}
\institutes{
$^{1}$ Department of Biomedical Engineering, University of Texas at Austin, Austin, TX, USA \\
$^{2}$ Department of Computer Science, University of Texas at Austin, Austin, TX, USA \\
$^{3}$ School of Information, University of Texas at Austin, Austin, TX, USA \\
$^{4}$ College of Natural Sciences, University of Texas at Austin, Austin, TX, USA \\
$^{5}$ Graphen, Inc., New York, NY, USA \\
$^{6}$ Department of Pediatrics, University of Texas at Austin, Austin, TX, USA \\
$^{7}$ Department of Cardiac Surgery, Division of Pediatric Cardiac Surgery, Vanderbilt University School of Medicine, Nashville, TN, USA \\
$^{8}$ Pediatric Heart Institute, Monroe Carell Jr. Children's Hospital at Vanderbilt, Nashville, TN, USA \\
$^{11}$ Department of Population Health, Dell Medical School, University of Texas at Austin, Austin, TX, USA \\
}
\maketitle
\section*{Abstract}
\textit{Thematic analysis (TA) is widely used in health research to extract patterns from patient interviews, yet manual TA faces challenges in scalability and reproducibility.
LLM-based automation can help, but existing approaches produce codebooks with limited generalizability and lack analytic auditability.
We present an automated TA framework combining iterative codebook refinement with full provenance tracking.
Evaluated on five corpora spanning clinical interviews, social media, and public transcripts, the framework achieves the highest composite quality score on four of five datasets compared to six baselines.
Iterative refinement yields statistically significant improvements on four datasets ($p < 0.01$, paired $t$-test, $n{=}5$ replicates) with large effect sizes ($d > 2.7$), driven by gains in code reusability and distributional consistency while preserving descriptive quality.
On two clinical corpora (pediatric cardiology), generated themes align with expert-annotated themes (mean cosine similarity .487--.494).
}
\section{Introduction}
Thematic analysis (TA) is essential for understanding patient and family experiences in health research~\cite{Braun2014, braun_clarke2021_thematicanalysis, Campbell2021}.
In pediatric cardiology, for example, interviews with families affected by congenital heart disease generate rich narratives~\cite{mery2023examining} that, when coded and organized into themes, yield actionable insights for care delivery.
However, manual TA faces well-documented challenges in scalability~\cite{Milat2012}, consistency~\cite{Vaismoradi2013}, and reproducibility~\cite{Braun2024}, particularly as clinical qualitative datasets grow.
Large language models (LLMs) have emerged as promising tools for automating this workflow, with early results demonstrating gains in efficiency and consistency~\cite{bijker2024chatgpt, dunivin2025scaling, mathis2024inductive}.
Yet a key gap limits their adoption in clinical settings: \textbf{generalizability gap}.
Single-pass coding methods produce codebooks that overfit to text seen during generation and fail to transfer to unseen data~\cite{pi2026logosllmdrivenendtoendgrounded}, a problem compounded in health research where codebooks must generalize across diverse patient narratives to support clinical guidelines and quality improvement initiatives~\cite{Braun2014, yi2025positionthematicanalysisunstructured}.
Furthermore, existing LLM-based frameworks report final themes without exposing intermediate decisions~\cite{pi2026logosllmdrivenendtoendgrounded, Thematic-LM, sharma2025details, Lam_2024, zhong2025hicodehierarchicalinductivecoding}, making it difficult for researchers to verify or reproduce the analytic process.

We present an automated TA framework that addresses the generalizability gap through an \textbf{iterative codebook refinement loop}, which progressively improves codebook generalizability by exposing it to diverse training samples across multiple rounds, consolidating overlapping codes and surfacing missing ones without degrading descriptive quality.
To address the auditability concern, the framework maintains \textbf{full provenance}: every artifact (quotes, codes, subthemes, and themes) receives a persistent identifier and every agent operation is logged in an auditable action ledger.
The pipeline integrates a grounded coding module~\cite{pi2026logosllmdrivenendtoendgrounded} with a synthesis stage (Auto-TA~\cite{yi2025autota}) that constructs the thematic hierarchy.
We evaluate on five corpora spanning clinical interviews (AAOCA, SV-CHD), social media (Dreaddit), research interviews (Sheffield), and public transcripts (Ali Abdaal), addressing three research questions:
\begin{enumerate}
    \item How does the framework's coding quality compare to existing automated methods across diverse datasets?
    \item Does iterative refinement produce statistically significant improvements in codebook generalizability?
    \item How well do generated themes align with human-annotated themes on clinical datasets?
\end{enumerate}
Across five datasets and six baselines, the framework achieves the highest composite quality score on four of five corpora, with iterative refinement yielding statistically significant improvements ($p < 0.01$, $n{=}5$ replicates) and large effect sizes ($d > 2.7$) on four datasets.
On the two clinical corpora, generated themes show meaningful alignment with expert-annotated themes (mean cosine similarity .487--.494).

\section{Related Work}
\subparagraph{Thematic Analysis in Health Research.}
TA is among the most widely adopted qualitative methods in health sciences, used to surface patient perspectives and unmet needs from interview and focus-group data~\cite{Braun2014, Campbell2021}.
In clinical settings, TA transforms unstructured narratives into themes that inform care delivery and policy.
Despite its value, TA remains labor-intensive: manual coding requires multiple trained analysts, iterative consensus-building, and extensive documentation~\cite{braun_clarke2021_thematicanalysis, Braun2024}, limiting the scale at which qualitative insights can be extracted from growing clinical data sources.
Established qualitative data analysis platforms (e.g., NVivo, Dedoose) support code-to-source traceability for manual workflows; however, existing LLM-based frameworks lack comparable provenance mechanisms.

\subparagraph{LLM-Assisted Qualitative Analysis.}
Recent work explores LLMs as qualitative coding assistants, from first-pass open coding~\cite{bijker2024chatgpt, zhang2024qualigpt, dunivin2025scaling} to end-to-end TA pipelines~\cite{yi2025sfttasupervisedfinetunedagents}.
Early approaches position LLMs as coding aides that generate candidate labels for researcher review~\cite{bijker2024chatgpt, mathis2024inductive}.
More recent systems automate larger portions of the workflow: concept induction with human-in-the-loop sensemaking (LLooM~\cite{Lam_2024}), hierarchical inductive coding (HICode~\cite{zhong2025hicodehierarchicalinductivecoding}), multi-agent workflows (Thematic-LM~\cite{Thematic-LM}), and iterative schema induction with generation--critique cycles (LOGOS~\cite{pi2026logosllmdrivenendtoendgrounded}).
Retrieval-augmented approaches (GraphRAG~\cite{edge2025localglobalgraphrag}, LightRAG~\cite{guo2025lightragsimplefastretrievalaugmented}) improve factual grounding, while DeTAILS~\cite{sharma2025details} and TAMA~\cite{xu2025tamahumanaicollaborativethematic} emphasize human-in-the-loop refinement.
A common limitation is that these systems diverge into two camps: some (LOGOS, HICode, LLooM) treat the codebook as the final output without synthesizing higher-order themes, while others (Thematic-LM, DeTAILS) generate themes directly but without an explicit codebook stage, sacrificing the code-level traceability needed for auditability. Our framework bridges this gap by combining iterative codebook refinement (via LOGOS) with inductive theme synthesis (via Auto-TA), producing a complete thematic hierarchy with end-to-end provenance from themes back to source evidence.
\section{Methods}
\paragraph{Datasets.}
We evaluate on five corpora spanning public and restricted-access qualitative data across clinical and non-clinical settings.
\textbf{(1) AAOCA transcripts.} Nine de-identified transcripts from nine focus groups with 42 parents of children with Anomalous Aortic Origin of a Coronary Artery~\cite{mery2023examining} (mean 10{,}987 words; collected Aug 2021--May 2022).
\textbf{(2) SV-CHD transcripts.} 28 de-identified transcripts from experience-group sessions and one-on-one interviews with patients, parents, siblings, and partners of families affected by Single-Ventricle Congenital Heart Disease~\cite{mery2023examining} (mean 11{,}198 words; collected Feb--Sep 2020). Approved by UT IRB (\#2019080031); available upon reasonable request.
\textbf{(3) Ali Abdaal transcripts.} We collected 163 YouTube video transcripts on productivity, entrepreneurship, and personal development (mean 6{,}477 words each), filtering out short-form content under 500 words.
\textbf{(4) Sheffield transcripts.} A publicly available collection of 15 semi-structured interviews (mean 6{,}927 words) with academic researchers discussing open research data sharing practices, hosted on the University of Sheffield's ORDA repository.
\textbf{(5) Dreaddit dataset.} A public corpus of 3{,}553 Reddit posts (~318{,}000 words) from ten stress-related subreddits~\cite{turcan-mckeown-2019-dreaddit}, included to test our framework generalization to informal, user-generated social-media text.
\subparagraph{The Traceable Thematic Analysis Framework.}
This framework provides an end-to-end automated thematic analysis pipeline with five main stages. Given a raw transcript, it produces a hierarchical thematic structure consisting of \emph{themes}, \emph{subthemes}, and \emph{codes}, each grounded in supporting \emph{quotes} from the source text. The analytical progression follows the conventional qualitative workflow: \emph{quotes} $\rightarrow$ \emph{codes} $\rightarrow$ \emph{subthemes} $\rightarrow$ \emph{themes}. The \emph{quotes} $\rightarrow$ \emph{codes} step is implemented by adopting the coding procedure of LOGOS~\cite{pi2026logosllmdrivenendtoendgrounded}, while the subsequent synthesis steps (codes $\rightarrow$ subthemes $\rightarrow$ themes) are performed by our Auto-TA~\cite{yi2025autota} module. A central design requirement of this framework is transparency: every intermediate artifact is assigned a persistent unique identifier, and every agent operation is recorded as an auditable action with a unique action ID, enabling full traceability from any final theme back to the exact transcript evidence. Details for each step are provided below.
\begin{figure}[h!]
\centering
\includegraphics[width=1\linewidth]{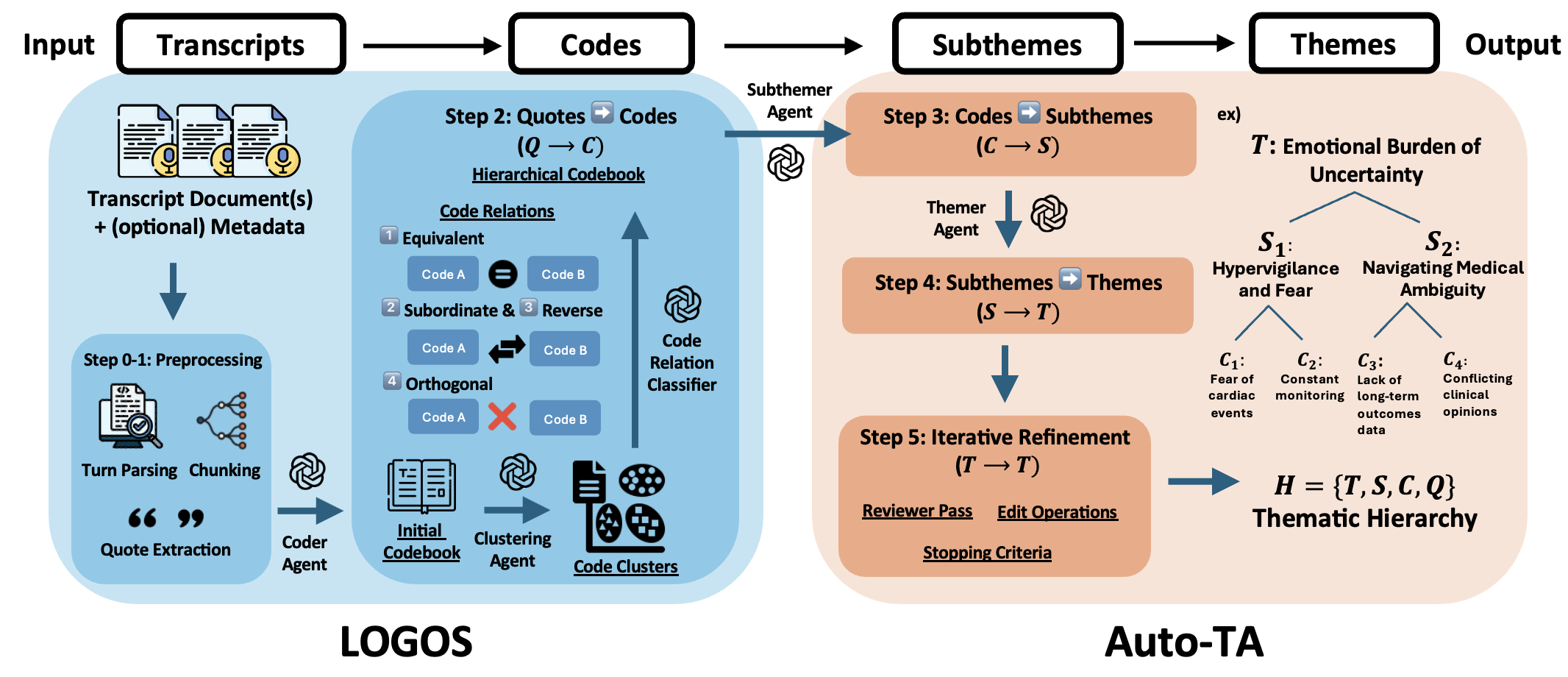}
\caption{The Traceable Thematic Analysis Framework}
\label{fig:traceable-ta}
\end{figure}
\paragraph{Inputs, outputs, and core artifacts.}
The input is a single transcript document (or a set of transcripts), represented as raw text with optional structural metadata (e.g., speaker labels, turn boundaries).
The output is a thematic hierarchy $\mathcal{H} = \{\mathcal{T}, \mathcal{S}, \mathcal{C}, \mathcal{Q}\}$ where themes $\mathcal{T}$ contain subthemes $\mathcal{S}$, subthemes contain codes $\mathcal{C}$, and codes link to evidence quotes $\mathcal{Q}$.
Each artifact type---quote, code, subtheme, and theme---stores a persistent unique identifier, a label or text span, a definition or description, and directed links to its parent and child artifacts, enabling full provenance traversal from any theme down to the exact transcript evidence.
\paragraph{Stages 0--1: Preprocessing, segmentation, and quote extraction.}
The framework first normalizes transcripts through \textit{turn parsing}, which segments text into speaker-attributed units with unique \texttt{turn\_id}s, and \textit{chunking}, which splits transcripts exceeding the model's context window into overlapping segments (default 8{,}000 characters) that respect natural boundaries. Each chunk receives a stable \texttt{chunk\_id} for cross-chunk consistency. During coding (Stage 2), the Coder agent simultaneously extracts candidate evidence quotes---short, self-contained passages conveying a single idea or experience---and assigns each a stable \texttt{quote\_id} linked to its source turn, enabling traceability back to the original transcript. Minimum length constraints filter non-substantive utterances, and passages exceeding 1{,}000 characters are automatically segmented. The output is a set of quote-code associations grounding every code in specific transcript evidence.
\paragraph{Stage 2: Grounded quotes $\rightarrow$ codes mapping (Coding Module).}
This stage assigns codes to transcript chunks via a modular coding backend that supports multiple instantiations. The default implementation follows the grounded coding procedure described in LOGOS~\cite{pi2026logosllmdrivenendtoendgrounded} : for each chunk, the coding agent generates a fixed number of codes (default 20), where each code comprises a short label (5--12 words) capturing the central meaning and a description (40--80 words) clarifying its scope. Codes are tagged with their source chunk identifier for traceability. After initial generation, a normalization pass deduplicates codes by name across chunks (incrementing frequency counts for repeated codes), then consolidates codes through pairwise relationship classification. For each pair of codes exceeding a cosine-similarity threshold, the LLM classifies the relationship as one of four types: \textit{equivalent} (same concept, merged via union-find), \textit{subordinate} (A is a subcategory of B), \textit{reverse} (B is a subcategory of A), or \textit{orthogonal} (unrelated). The classified relations are assembled into a directed code hierarchy graph with transitive inference (e.g., if A$\rightarrow$B and B$\rightarrow$C, then A$\rightarrow$C). A three-step cleanup then applies: (1)~equivalent codes are merged, retaining the code with the highest merge score (weighted by frequency and in-degree); (2)~low-frequency codes with parents are subsumed into their parent; (3)~remaining low-frequency orphan codes are dropped. Alternative coding backends~\cite{strauss_corbin_1998_basics, Lam_2024, zhong2025hicodehierarchicalinductivecoding, edge2025localglobalgraphrag, guo2025lightragsimplefastretrievalaugmented} can be substituted at this stage for systematic comparison across methods (see Table~\ref{tab:performance}). The coding stage outputs the code set $\mathcal{C}$ with frequency and hierarchy metadata.
\paragraph{Stages 3--4: Auto-TA~\cite{yi2025autota} synthesis (codes $\rightarrow$ subthemes $\rightarrow$ themes).}
Auto-TA constructs a two-level thematic hierarchy through two successive passes. In the first pass, semantically related codes are grouped into candidate subthemes $\mathcal{S}$ based on shared meaning or complementary facets, generating for each a subtheme label, a description capturing the common thread, and the list of constituent codes. In the second pass, related subthemes are aggregated into overarching themes $\mathcal{T}$ by identifying higher-order patterns across subtheme clusters, generating for each a theme label (5--10 words), a detailed description (60--80 words), and the list of subsumed subthemes. The system ensures comprehensive coverage by assigning every code to at least one subtheme and every subtheme to at least one theme, while maintaining semantic distinctness across both hierarchical levels.
\paragraph{Stage 5: Iterative refinement and stabilization.}
This stage refines the hierarchy over multiple rounds, typically alternating between (i) structural edits and (ii) semantic edits.
\begin{itemize}
    \item \textbf{Reviewer pass.} A reviewer agent checks for common failure modes such as: duplicated concepts across branches, inconsistent granularity (some themes too broad while others too narrow), orphan codes/subthemes, and weak grounding (themes supported by very few quotes).
    \item \textbf{Edit operations.} Refinements are expressed using a constrained set of actions:
    \texttt{generate} (create new artifact),
    \texttt{merge} (combine artifacts),
    \texttt{split} (divide an artifact),
    \texttt{revise} (edit label/definition),
    \texttt{move} (reassign child links),
    \texttt{delete} (remove artifact).
    Each action produces a new action entry $a \in \mathcal{A}$ with explicit provenance.
    \item \textbf{Stopping criteria.} Refinement stops when the reviewer proposes no substantive structural edits in a full pass (or when a configured maximum number of rounds is reached). The final output is the stabilized hierarchy $\mathcal{H}$.
\end{itemize}
\paragraph{Transparency via provenance and action logging.}
To ensure transparency, we maintain an action ledger $\mathcal{A}$ where each agent operation is logged as:
\[
\mathcal{A} = \langle \texttt{aid}, \texttt{agent\_role}, \texttt{action\_type}, \texttt{inputs}, \texttt{outputs}, \texttt{justification}, \texttt{timestamp} \rangle.
\]
The \texttt{inputs}/\texttt{outputs} fields store the specific artifact IDs affected (e.g., merging codes \texttt{cid\_1,cid\_2} into \texttt{cid\_new}). This makes the pipeline \emph{replayable}: users can reconstruct how any theme evolved across iterations and inspect exactly which operations produced (or altered) each artifact. For example, a generated theme such as ``Communication challenges in healthcare'' can be traced through its constituent subthemes and codes down to specific evidence quotes (e.g., \texttt{Q47} from Turn~15 of Transcript~3), with every grouping and refinement step recorded in $\mathcal{A}$.
\paragraph{Experiments.} We address three research questions: (1) How does our framework compare to other coding methods across diverse datasets? (2) Does iterative refinement in the coding stage improve performance? (3) How do the generated themes align with human-annotated ground truth?
\subparagraph{Experiment 1 -- Comparative Performance (RQ1).}
This experiment evaluates whether LOGOS produces codebooks of higher quality than existing automated coding methods.
All methods are evaluated on the same train/test splits using the five quality metrics described in Section~\ref{sec:evaluation-metrics}; LLM-based metrics (\textcolor[HTML]{ff7f0e}{\textbf{Fitness}}, \textcolor[HTML]{2ca02c}{\textbf{Coverage}}) use GPT-4o-mini at temperature~$0.3$ on 5 sampled test chunks.
All baselines share the same underlying LLM to isolate algorithmic differences from model capacity.
We report two LOGOS variants: \textbf{LOGOS (Iter-1)}, the codebook from the initial pipeline before refinement, and \textbf{LOGOS (Best)}, the codebook from the iteration achieving the highest composite score.
To assess robustness, we run $n{=}5$ independent replicates per dataset (seeds $s \in \{42{-}46\}$, up to 10 iterations each) with the train/test split held fixed, and report mean scores with significance via two-tailed paired $t$-tests (df\,=\,4) and Cohen's $d$.
Refinement terminates early when the Jaccard similarity between successive codebooks exceeds $0.95$.
\subparagraph{Experiment 2 -- Iterative Refinement (RQ2).}
This experiment isolates the refinement loop's contribution by tracking all five metrics across iterations.
For each dataset, LOGOS produces an initial codebook (Iter-0), then refines it for up to 10 iterations---sampling training chunks, proposing code revisions, and re-evaluating on the held-out test set. Statistical testing follows Experiment~1 (paired $t$-tests, Cohen's $d$, $n{=}5$).
\subparagraph{Experiment 3 -- Theme Alignment with Human Annotations (RQ3).}
For the two clinical datasets (AAOCA and SV-CHD), we compare generated themes against independently developed human-annotated themes~\cite{mery2023examining} by computing cosine similarity (all-MiniLM-L6-v2) between each generated theme and all human themes, reporting the closest match as a quantitative proxy for semantic alignment.
\paragraph{Evaluation Metrics.}
\label{sec:evaluation-metrics}
We adopt the same evaluation protocol and codebook-quality metrics as in previous work~\cite{pi2026logosllmdrivenendtoendgrounded} . Concretely, the codebook is learned on the training split, and we then perform deductive coding on the test split using only the train-derived codebook (i.e., the model selects up to 20 existing codes and is not allowed to introduce new codes). We report the following metrics, computed from the resulting test-set code assignments.
\begin{itemize}
    \item \textcolor[HTML]{1f77b4}{\textbf{Reusability}} measures how many codes in the learned codebook are actually used on unseen data:
    \begin{equation}
        \mathrm{Reusability} = \frac{\#\text{used codes}}{\#\text{all codes}},
    \end{equation}
    where a \emph{used} code is assigned at least once to a test chunk.
    \item Descriptive  \textcolor[HTML]{ff7f0e}{\textbf{Fitness}} is an LLM-judged score (originally 1--10, rescaled to $[0,1]$) that quantifies how well the assigned codes appropriately describe a test chunk (even if only partially).
    \item Descriptive \textcolor[HTML]{2ca02c}{\textbf{Coverage}} is an LLM-judged score (originally 1--10, rescaled to $[0,1]$) that evaluates whether the assigned codes capture all essential aspects of a test chunk relevant to the research question (i.e., completeness).
    \item \textcolor[HTML]{d62728}{\textbf{Parsimoniousness}} captures redundancy among codes via average pairwise cosine similarity:
    \begin{equation}
        \mathrm{Parsimoniousness} = 1 - \frac{2}{n(n-1)} \sum_{i<j} \cos(\mathbf{c}_i, \mathbf{c}_j),
    \end{equation}
    where $n$ is the number of codes and $\cos(\mathbf{c}_i, \mathbf{c}_j)$ is cosine similarity between code representations.   \item \textcolor[HTML]{9467bd}{\textbf{Consistency}} measures train--test agreement in code usage by penalizing divergence between the empirical code distributions:
    \begin{equation}
        \mathrm{Consistency} = 1 - \mathrm{JSD}(P \,\|\, Q),
    \end{equation}
    where $P$ and $Q$ are the empirical code-frequency distributions on the training and test splits, respectively.
    \item \textbf{Composite score} combines the above dimensions via a weighted sum; following LOGOS~\cite{pi2026logosllmdrivenendtoendgrounded} , we use equal weights unless otherwise stated.
\end{itemize}
\subparagraph{Experimental Settings.}
All experiments use \texttt{gpt-4o-mini} via the OpenAI API; each corpus is segmented into 2{,}048-word chunks with 200-word overlap, split into training (80\%) and test (20\%) sets (seed $s{=}42$).
Embedding-based metrics use \texttt{all-MiniLM-L6-v2} (384-d). All baselines share the same LLM, chunking, and splits.
\subsection{Results and Discussion}
\begin{figure}[h!]
\centering
\includegraphics[width=0.6\linewidth]{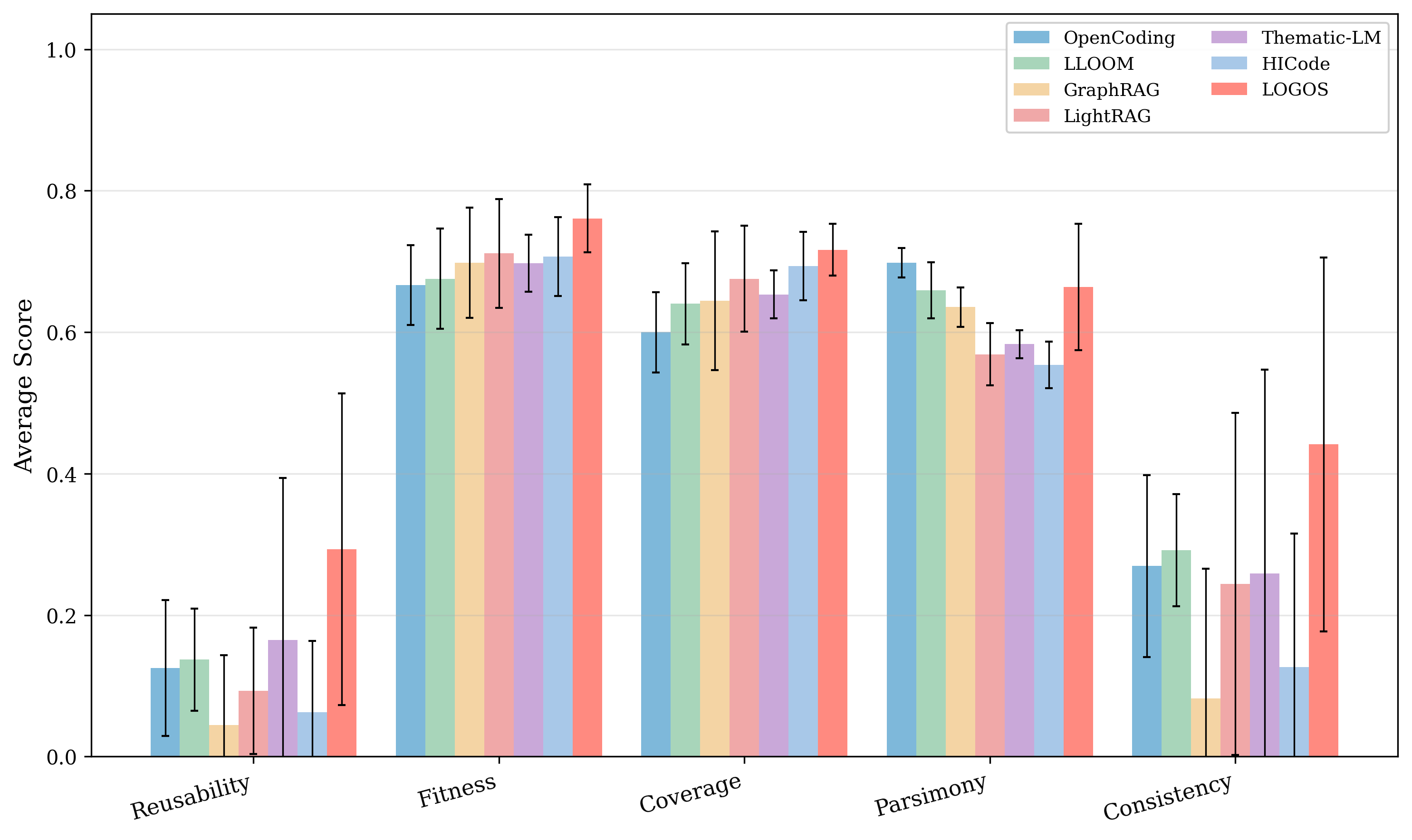}
\caption{Average quality metric scores across five datasets. Error bars indicate $\pm \sigma$  across datasets.}
\label{fig:metrics_breakdown}
\end{figure}

\begin{figure*}[t]
  \centering
  \includegraphics[width=\textwidth]{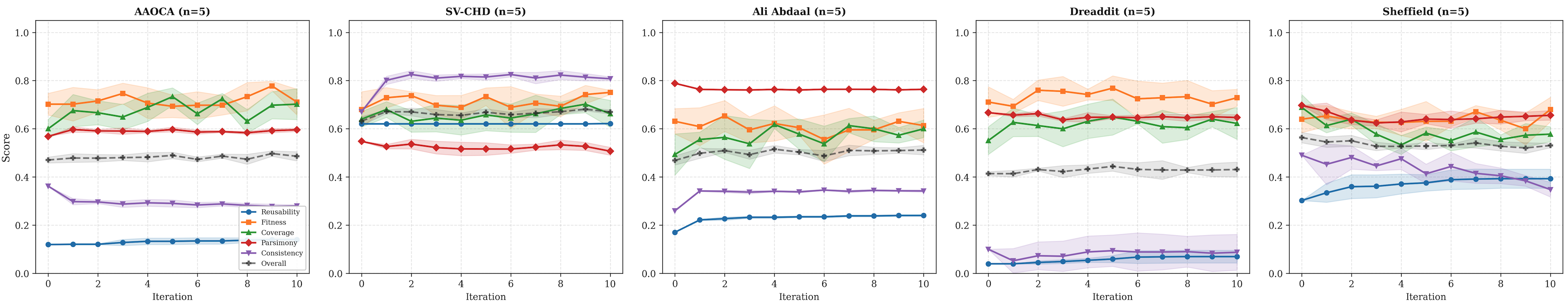}
  \caption{Quality metric trajectories across iterative refinement for all five datasets ($n{=}5$ replicates each). Solid lines show the mean; shaded bands denote 95\% confidence intervals. Metrics shown: \textcolor[HTML]{1f77b4}{\textbf{Reusability}}, \textcolor[HTML]{ff7f0e}{\textbf{Fitness}}, \textcolor[HTML]{2ca02c}{\textbf{Coverage}}, \textcolor[HTML]{d62728}{\textbf{Parsimony}}, \textcolor[HTML]{9467bd}{\textbf{Consistency}}, and \textbf{Overall} (dashed grey; unweighted mean). Iteration 0 is the initial codebook before refinement. Gains emerge within 2--4 iterations and plateau thereafter.}
  \label{fig:metrics_grid}
\end{figure*}

\subparagraph{Experiment 1 -- Comparative Performance (RQ1).}
We evaluated LOGOS against six baseline methods across five datasets (Table~\ref{tab:performance}). LOGOS (Best) achieves the highest composite score on four of five datasets, with statistically significant improvements ($p < 0.01$, paired $t$-test) over the initial codebook on Ali ($+.064$), Dreaddit ($+.048$), AAOCA ($+.039$), and SV-CHD ($+.056$). The largest absolute score is on SV-CHD (0.688), outperforming the next-best method by $+.187$. On Sheffield, Thematic-LM~\cite{Thematic-LM} achieves the highest score (0.629); LOGOS reaches 0.571 with a non-significant improvement ($+.007$, $p = 0.13$), likely due to the small corpus size limiting refinement diversity.
Figure~\ref{fig:metrics_breakdown} presents the per-metric breakdown. LOGOS (Best) achieves the highest \textbf{\textcolor[HTML]{1f77b4}{Reusability}} (0.41) and \textbf{\textcolor[HTML]{9467bd}{Consistency}} (0.57) by substantial margins, indicating superior generalizability, with trade-offs in \textbf{\textcolor[HTML]{ff7f0e}{Fitness}} (0.62) and \textbf{\textcolor[HTML]{2ca02c}{Coverage}} (0.60) where single-pass baselines that optimize for immediate descriptive quality score higher. The composite scores demonstrate that generalizability gains outweigh modest decreases in descriptive metrics, yielding superior overall performance for codebooks intended for use beyond the training corpus. A detailed per-metric statistical breakdown is provided in Appendix~\ref{app:statistical}.
\begin{table*}[t]
\centering
\caption{Comparative performance across methods. Scores report the composite quality metric (mean of Reusability, Fitness, Coverage, Parsimony, and Consistency). LOGOS (Iter-1) and LOGOS (Best) report means over $n{=}5$ independent replicates. $\Delta$ denotes improvement from iterative refinement. Statistical significance is assessed via paired $t$-tests: 
%\textbf{**}\,$p<0.05$; 
\textbf{***}\,$p<0.01$.}
\label{tab:performance}
\small
\begin{tabular}{l ccccc ccccc}
\toprule
& \multicolumn{2}{c}{\textbf{AAOCA~\cite{mery2023examining}}}
& \multicolumn{2}{c}{\textbf{SV-CHD~\cite{mery2023examining}}}
& \multicolumn{2}{c}{\textbf{Ali}}
& \multicolumn{2}{c}{\textbf{Dreaddit~\cite{turcan-mckeown-2019-dreaddit}}}
& \multicolumn{2}{c}{\textbf{Sheffield}} \\
\textbf{Method} 
& Score & $\Delta$ 
& Score & $\Delta$ 
& Score & $\Delta$ 
& Score & $\Delta$ 
& Score & $\Delta$ \\
\midrule
OpenCoding~\cite{strauss_corbin_1998_basics}      
& 0.493 & 
& 0.450 & 
& 0.473 & 
& 0.428 & 
& 0.516 & \\
LLOOM~\cite{Lam_2024}           
& 0.476 & 
& 0.495 & 
& 0.491 & 
& \textbf{0.458} & 
& 0.484 & \\
GraphRAG~\cite{edge2025localglobalgraphrag}        
& 0.433 & 
& 0.391 & 
& 0.431 & 
& 0.382 & 
& 0.467 & \\
LightRAG~\cite{guo2025lightragsimplefastretrievalaugmented}        
& 0.415 & 
& 0.501 & 
& 0.504 & 
& 0.368 & 
& 0.505 & \\
Thematic-LM~\cite{Thematic-LM}     
& 0.485 & 
& 0.473 & 
& 0.395 & 
& 0.375 & 
& \textbf{0.629} & \\
HICode~\cite{zhong2025hicodehierarchicalinductivecoding}          
& 0.412 & 
& 0.378 & 
& 0.421 & 
& 0.436 & 
& 0.496 & \\
LOGOS~\cite{pi2026logosllmdrivenendtoendgrounded} (Iter-1)  
& 0.471 & 
& 0.632 & 
& 0.469 & 
& 0.414 & 
& 0.564 & \\
\rowcolor{purple!15}
LOGOS~\cite{pi2026logosllmdrivenendtoendgrounded} (Best)    
& \textbf{0.509} & \textcolor{teal}{+.039}$^{***}$ 
& \textbf{0.688} & \textcolor{teal}{+.056}$^{***}$ 
& \textbf{0.533} & \textcolor{teal}{+.064}$^{***}$ 
& 0.462 & \textcolor{teal}{+.048}$^{***}$ 
& 0.571 & \textcolor{teal}{+.007} \\
\bottomrule
\end{tabular}
\end{table*}
\subparagraph{Experiment 2 -- Iterative Refinement (RQ2).}
To understand the dynamics of iterative refinement, we tracked all five quality metrics across 10 iterations on each dataset (Figure~\ref{fig:metrics_grid}).
The most striking pattern is the behavior of \textbf{\textcolor[HTML]{1f77b4}{Reusability}}: starting from low initial values (0.05--0.20), it shows consistent upward trends across all datasets, roughly doubling on AAOCA and SV-CHD by iteration 10. In contrast, \textbf{\textcolor[HTML]{ff7f0e}{Fitness}} and \textbf{\textcolor[HTML]{2ca02c}{Coverage}} remain stable (0.60--0.75) throughout refinement, confirming that descriptive quality is established early and preserved. \textbf{\textcolor[HTML]{d62728}{Parsimony}} is similarly stable (0.55--0.75), while \textbf{\textcolor[HTML]{9467bd}{Consistency}} shows dataset-dependent variability---rising sharply on Sheffield but remaining flat on Dreaddit. The composite \textbf{Overall} score improves gradually, driven primarily by \textbf{\textcolor[HTML]{1f77b4}{Reusability}} gains, with convergence typically occurring within 2--4 iterations on AAOCA and SV-CHD and continuing through iteration 10 on Ali Abdaal and Sheffield.
These trajectories confirm that iterative refinement improves generalizability (\textbf{\textcolor[HTML]{1f77b4}{Reusability}}, \textbf{\textcolor[HTML]{9467bd}{Consistency}}) without degrading descriptive quality (\textbf{\textcolor[HTML]{ff7f0e}{Fitness}}, \textbf{\textcolor[HTML]{2ca02c}{Coverage}}, \textbf{\textcolor[HTML]{d62728}{Parsimony}}), and that the gains in Table~\ref{tab:performance} emerge from cumulative refinement rather than initial codebook generation alone.
\subparagraph{Experiment 3 -- Theme Alignment with Human Annotations (RQ3).}
To assess clinical meaningfulness, we compared generated themes against human-annotated ground truth~\cite{mery2023examining} using cosine similarity (all-MiniLM-L6-v2; Table~\ref{tab:theme_comparison}). On AAOCA (8 themes, mean similarity .487) and SV-CHD (13 themes, mean .494), the framework recovers themes that align with core human-identified constructs.
The strongest alignments ($>.5$) capture the most emotionally salient family experiences: communication breakdowns between families and providers (.726), evolving parenting roles during treatment transitions (.620), and parental protective instincts around child safety (.587). These reflect central concerns in the pediatric cardiology qualitative literature~\cite{mery2023examining}.
Moderate-alignment themes (.4--.5) identify similar constructs but frame them from different analytical angles (e.g., ``Parental advocacy \& empowerment'' (.452) emphasizes family agency, while the matched human theme foregrounds information processing). Lower-alignment themes ($<.4$) surface dimensions not prioritized in the original manual analysis or group more broadly than human coders.
A notable pattern across all tiers is that generated themes tend toward higher abstraction than human themes: for instance, ``Parenting \& family dynamics'' is broad enough to encompass many subconcepts, whereas the matched human theme ``Change in parent role as child transitions'' is specific and clinically actionable. This suggests that without domain-specific constraints, the framework favors generality over the targeted specificity that expert coders bring. Additionally, cosine similarity based on sentence embeddings can overestimate alignment when themes share topical vocabulary but differ in meaning---e.g., ``Balancing caregiving \& professional responsibilities'' (.558) and ``Partnership with the care team'' address distinct constructs despite lexical overlap in the caregiving domain.
To illustrate the framework's traceability, consider the generated AAOCA theme ``Emotional and psychological impacts'': it decomposes into subthemes such as ``Need for counseling support,'' which groups codes like ``Proactive information acquisition for child health decisions,'' each grounded in participant quotes (e.g., P4006: ``the therapy, the counseling should be mandatory for the child or for the family''). Every link in this chain---theme to subtheme to code to quote to transcript turn---is recorded in the action ledger, enabling researchers to audit how any theme was derived from source evidence.
Formal evaluation with domain experts remains necessary to validate true semantic alignment and determine whether divergences represent complementary perspectives or missed clinical nuances.
\begin{table}[t]
\centering
\caption{Comparison of framework-generated themes with human-annotated themes~\cite{mery2023examining}, matched by cosine similarity (all-MiniLM-L6-v2). Rows sorted by descending similarity.}
\label{tab:theme_comparison}
\small
\setlength{\tabcolsep}{3pt}
\renewcommand{\arraystretch}{0.95}
\begin{tabular}{@{} l l r @{}}
\toprule
\textbf{Generated Theme} & \textbf{Closest Human Theme} & \textbf{Sim.} \\
\midrule
\multicolumn{3}{@{}l}{\textbf{AAOCA} \textit{(8 generated; mean = .487)}} \\
\midrule
Parental fears \& protective instincts & Feeling that my child is safe & .587 \\
Balancing caregiving \& prof.\ responsibilities & Partnership with the care team & .558 \\
Child's engagement \& understanding & Feeling that my child is safe & .515 \\
Emotional \& psychological impacts & Coping with stress, anxiety, depression & .504 \\
Communication complexities in health mgmt.\ & Communicating risks appropriately & .496 \\
Healthcare communication \& trust & Empathetic relationship-centered care & .480 \\
Managing asymptomatic \& chronic conditions & No consistent diagnostic/treatment paradigm & .446 \\
Post-operative recovery \& normalization & No consistent diagnostic/treatment paradigm & .308 \\
\midrule
\multicolumn{3}{@{}l}{\textbf{SV-CHD} \textit{(13 generated; mean = .494)}} \\
\midrule
Communication challenges in healthcare & Communication deficiencies disrupt continuity & .726 \\
Parenting \& family dynamics & Change in parent role as child transitions & .620 \\
Emotional experiences of parents & Rarely receive specialized mental health care & .558 \\
Hospitalization \& recovery experiences & Hospital-to-home transition presents challenges & .547 \\
Healthcare resource \& expertise challenges & Other specialties have limited CHD knowledge & .525 \\
Children's understanding \& involvement & Change in parent role as child transitions & .493 \\
Navigating healthcare systems & Hospital-to-home transition presents challenges & .469 \\
Adapting to lifestyle changes & Smooth transition between life stages & .463 \\
Parental advocacy \& empowerment & Parents synthesize critical medical information & .452 \\
Decision-making \& autonomy & Accessible information about goals/expectations & .421 \\
Medical interventions \& their impact & Preparation to independently manage condition & .402 \\
Educational \& informational needs & Accessible information about goals/expectations & .380 \\
Support networks \& community & Lack connection to peers affected by SV-CHD & .370 \\
\bottomrule
\end{tabular}
\end{table}
\section{Conclusion, Limitations, and Future Work}
We presented a traceable thematic analysis framework with end-to-end provenance from themes back to source evidence. Across five datasets and six baselines, iterative refinement produces statistically significant improvements ($p < 0.01$, $n{=}5$) with large effect sizes ($d > 2.7$) on four datasets, selectively improving generalizability (\textbf{\textcolor[HTML]{1f77b4}{Reusability}}, \textbf{\textcolor[HTML]{9467bd}{Consistency}}) while preserving descriptive quality.
Several limitations remain. First, the optimal refinement iteration is identified post hoc; a principled early-stopping criterion beyond the Jaccard proxy is needed. Second, the co-movement of \textbf{\textcolor[HTML]{1f77b4}{Reusability}} and \textbf{\textcolor[HTML]{9467bd}{Consistency}} suggests these metrics may capture overlapping facets of generalizability, and Sheffield's non-significant improvement highlights a corpus-size dependency. Third, \textbf{\textcolor[HTML]{ff7f0e}{Fitness}} and \textbf{\textcolor[HTML]{2ca02c}{Coverage}} rely on LLM-based judgments from the same model family used for generation, potentially introducing systematic bias. Fourth, embedding-based cosine similarity can inflate theme alignment scores when themes share domain vocabulary but differ in meaning, and generated themes tend toward higher abstraction than clinically specific human-annotated themes---expert qualitative review and domain constraints could address both issues. Finally, integrating human-in-the-loop checkpoints for safety-critical healthcare applications and reducing per-replicate API cost are key future directions.
\section*{Appendix}
\appendix
\section{Statistical Analysis of Iterative Refinement}
\label{app:statistical}
To rigorously evaluate whether LOGOS's iterative refinement loop yields meaningful improvements, we conducted $n{=}5$ independent replicates of the full refinement pipeline (10 iterations each) on every dataset, varying the random seed that governs chunk sampling during both refinement and evaluation.
Table~\ref{tab:metric_breakdown} reports the per-metric breakdown comparing the initial codebook (Iter-1) against the best-performing iteration within each replicate.
Statistical significance is assessed via two-tailed paired $t$-tests on the per-replicate scores (df\,=\,4), and effect magnitude is quantified using Cohen's $d$ for paired samples.
Across all five datasets, \textcolor[HTML]{ff7f0e}{\textbf{Fitness}} and \textcolor[HTML]{2ca02c}{\textbf{Coverage}}---the two metrics that directly measure how well codes describe and capture the content of unseen text---exhibit the most consistent gains.
\textcolor[HTML]{2ca02c}{\textbf{Coverage}} improvements range from $+.013$ (Sheffield) to $+.173$ (Dreaddit), reaching significance ($p < 0.01$) on four of five datasets with uniformly large effect sizes ($d > 1.6$).
\textcolor[HTML]{ff7f0e}{\textbf{Fitness}} gains range from $+.071$ to $+.107$ and are significant on three datasets.
\textcolor[HTML]{9467bd}{\textbf{Consistency}}, which measures the stability of code-frequency distributions between train and test splits, shows the single largest per-metric improvement on SV-CHD ($+.170$, $p < 0.001$, $d = 18.99$), indicating that refinement substantially stabilizes how codes distribute across unseen data.
\textcolor[HTML]{1f77b4}{\textbf{Reusability}}---the fraction of codes that match at least one test chunk---improves significantly on Ali ($+.070$, $p < 0.001$), AAOCA ($+.021$, $p < 0.01$), and Sheffield ($+.090$, $p < 0.05$).
\textcolor[HTML]{d62728}{\textbf{Parsimony}} gains are modest, reaching significance only on AAOCA ($+.039$, $p < 0.001$); on datasets where the initial codebook is already compact, there is limited room for further redundancy reduction.
The composite \textbf{Overall} score improves significantly ($p < 0.01$) on four of five datasets, with large effect sizes ($d > 2.7$) on AAOCA, SV-CHD, Ali, and Dreaddit.
Sheffield is the sole exception ($\Delta = +.007$, $p = 0.13$), which we attribute to its small corpus (15 documents, 49 train chunks): the limited diversity of available text constrains the refinement loop's ability to discover meaningfully distinct codes across iterations.
Notably, even on Sheffield, all per-metric deltas are non-negative and the effect size for the overall score is still large ($d = 0.84$), suggesting that the lack of significance reflects low statistical power rather than a failure of the refinement mechanism.
Figure~\ref{fig:metrics_grid} visualizes the trajectory of each metric across iterations, with shaded bands denoting 95\% confidence intervals computed over the five replicates.
The plots confirm that improvements emerge within the first 2--4 iterations and generally plateau thereafter, consistent with the Jaccard-based convergence criterion (Section~\ref{sec:evaluation-metrics}).
\\
\begin{table*}[h]
\centering
\caption{Per-metric statistical analysis of iterative refinement (Iter-1 vs.\ Best). All values are means over $n{=}5$ independent replicates. $\Delta$: improvement from refinement. $t$: paired $t$-statistic (df\,=\,4). $d$: Cohen's $d$ (paired). \textbf{**}\,$p<0.05$; \textbf{***}\,$p<0.01$. Effect sizes: S\,=\,small ($|d|<0.5$), M\,=\,medium ($|d|<0.8$), L\,=\,large ($|d|\geq0.8$).}
\label{tab:metric_breakdown}
\small
\setlength{\tabcolsep}{4pt}
\begin{tabular}{ll rrr rr cc}
\toprule
\textbf{Dataset} & \textbf{Metric} & \textbf{Iter-1} & \textbf{Best} & $\boldsymbol{\Delta}$ & $\boldsymbol{t}$ & $\boldsymbol{p}$ & \textbf{Sig.} & $\boldsymbol{d}$\,/\,\textbf{Effect} \\
\midrule
AAOCA & Reusability  & .119 & .140 & +.021 & $-$4.98  & .008  & *** & 2.23\,/\,L \\
      & Fitness      & .702 & .791 & +.089 & $-$4.00  & .016  & **  & 1.79\,/\,L \\
      & Coverage     & .600 & .760 & +.160 & $-$6.47  & .003  & *** & 2.89\,/\,L \\
      & Parsimony    & .569 & .608 & +.039 & $-$15.92 & $<$.001 & *** & 7.12\,/\,L \\
      & Consistency  & .362 & .362 & .000  & ---      & ---   & --- & ---  \\
      & \cellcolor{gray!10}Overall & \cellcolor{gray!10}.471 & \cellcolor{gray!10}.509 & \cellcolor{gray!10}+.039 & \cellcolor{gray!10}$-$6.23 & \cellcolor{gray!10}.003 & \cellcolor{gray!10}*** & \cellcolor{gray!10}2.78\,/\,L \\
\midrule
SV-CHD & Reusability  & .620 & .622 & +.001 & $-$1.00  & .374  &     & 0.45\,/\,S \\
       & Fitness      & .680 & .773 & +.093 & $-$2.54  & .064  &     & 1.13\,/\,L \\
       & Coverage     & .640 & .733 & +.093 & $-$5.72  & .005  & *** & 2.56\,/\,L \\
       & Parsimony    & .548 & .555 & +.007 & $-$2.35  & .079  &     & 1.05\,/\,L \\
       & Consistency  & .671 & .841 & +.170 & $-$42.46 & $<$.001 & *** & 18.99\,/\,L \\
       & \cellcolor{gray!10}Overall & \cellcolor{gray!10}.632 & \cellcolor{gray!10}.688 & \cellcolor{gray!10}+.056 & \cellcolor{gray!10}$-$6.89 & \cellcolor{gray!10}.002 & \cellcolor{gray!10}*** & \cellcolor{gray!10}3.08\,/\,L \\
\midrule
Ali Abdaal & Reusability  & .171 & .240 & +.070 & $-$47.18 & $<$.001 & *** & 21.10\,/\,L \\
           & Fitness      & .631 & .711 & +.080 & $-$4.81  & .009  & *** & 2.15\,/\,L \\
           & Coverage     & .493 & .662 & +.169 & $-$3.73  & .020  & **  & 1.67\,/\,L \\
           & Parsimony    & .789 & .789 & .000  & ---      & ---   & --- & ---  \\
           & Consistency  & .260 & .347 & +.087 & $-$124.08 & $<$.001 & *** & 55.49\,/\,L \\
           & \cellcolor{gray!10}Overall & \cellcolor{gray!10}.469 & \cellcolor{gray!10}.533 & \cellcolor{gray!10}+.064 & \cellcolor{gray!10}$-$6.56 & \cellcolor{gray!10}.003 & \cellcolor{gray!10}*** & \cellcolor{gray!10}2.93\,/\,L \\
\midrule
Dreaddit & Reusability  & .040 & .070 & +.030 & $-$2.23  & .090  &     & 1.00\,/\,L \\
         & Fitness      & .711 & .818 & +.107 & $-$4.95  & .008  & *** & 2.21\,/\,L \\
         & Coverage     & .551 & .724 & +.173 & $-$6.29  & .003  & *** & 2.81\,/\,L \\
         & Parsimony    & .667 & .669 & +.002 & $-$1.56  & .195  &     & 0.70\,/\,M \\
         & Consistency  & .100 & .130 & +.030 & $-$1.17  & .307  &     & 0.52\,/\,M \\
         & \cellcolor{gray!10}Overall & \cellcolor{gray!10}.414 & \cellcolor{gray!10}.462 & \cellcolor{gray!10}+.048 & \cellcolor{gray!10}$-$6.33 & \cellcolor{gray!10}.003 & \cellcolor{gray!10}*** & \cellcolor{gray!10}2.83\,/\,L \\
\midrule
Sheffield & Reusability  & .302 & .393 & +.090 & $-$4.50  & .011  & **  & 2.01\,/\,L \\
          & Fitness      & .640 & .711 & +.071 & $-$1.61  & .182  &     & 0.72\,/\,M \\
          & Coverage     & .689 & .702 & +.013 & $-$1.00  & .374  &     & 0.45\,/\,S \\
          & Parsimony    & .697 & .700 & +.002 & $-$1.60  & .184  &     & 0.72\,/\,M \\
          & Consistency  & .490 & .527 & +.036 & $-$2.18  & .094  &     & 0.98\,/\,L \\
          & \cellcolor{gray!10}Overall & \cellcolor{gray!10}.564 & \cellcolor{gray!10}.571 & \cellcolor{gray!10}+.007 & \cellcolor{gray!10}$-$1.88 & \cellcolor{gray!10}.134 & \cellcolor{gray!10} & \cellcolor{gray!10}0.84\,/\,L \\
\bottomrule
\end{tabular}
\end{table*}
% References as numbers
\makeatletter
\renewcommand{\@biblabel}[1]{\hfill #1.}
\makeatother

{\small
\bibliographystyle{vancouver}
\bibliography{amia}
} 
\end{document}